# PARALLEL SWIN TRANSFORMER-ENHANCED 3D MRI-TO-CT SYNTHESIS FOR MRI-ONLY RADIOTHERAPY PLANNING

*Zolnamar Dorjsembe[1,3], Hung-Yi Chen[2*], Furen Xiao[2†,3], Hsing-Kuo Pao[1]*

[1] Department of Computer Science, National Taiwan University of Science and Technology, Taiwan
[2*] Department of Oncology, [†] Department of Surgery, National Taiwan University Hospital, Taiwan
[3] Institute of Medical Device and Imaging, College of Medicine, National Taiwan University, Taiwan

## ABSTRACT

MRI provides superior soft tissue contrast without ionizing radiation; however, the absence of electron density information limits its direct use for dose calculation. As a result, current radiotherapy workflows rely on combined MRI and CT acquisitions, increasing registration uncertainty and procedural complexity. Synthetic CT generation enables MRI only planning but remains challenging due to nonlinear MRI–CT relationships and anatomical variability. We propose Parallel Swin Transformer-Enhanced Med2Transformer, a 3D architecture that integrates convolutional encoding with dual Swin Transformer branches to model both local anatomical detail and long-range contextual dependencies. Multi-scale shifted window attention with hierarchical feature aggregation improves anatomical fidelity. Experiments on public and clinical datasets demonstrate higher image similarity and improved geometric accuracy compared with baseline methods. Dosimetric evaluation shows clinically acceptable performance, with a mean target dose error of 1.69%. Code is available at: https://github.com/mobaidoctor/med2transformer.

## 1. INTRODUCTION

Magnetic Resonance Imaging (MRI) provides superior soft-tissue contrast without ionizing radiation, making it invaluable for precise target and organ-at-risk delineation in radiotherapy. However, MRI lacks electron density information required for dose calculation, which remains dependent on Computed Tomography (CT). Consequently, current workflows rely on both MRI and CT, increasing cost, patient burden, and registration uncertainty [1].

To address this limitation, synthetic CT (sCT) generation from MRI has been proposed to enable MRI-only radiotherapy planning [2]. Deep learning has significantly advanced sCT generation, yet accurate synthesis remains challenging due to nonlinear MRI-CT intensity relationships, tissue heterogeneity, and scanner variability [1]. Moreover, generalizing across different anatomical sites and imaging protocols while maintaining geometric and dosimetric accuracy remains an open challenge [3].

Convolutional neural networks (CNNs) effectively capture local structures but struggle with global dependencies, while transformer-based models capture long-range context but often lose fine spatial details [4]. Bridging these complementary strengths is critical for accurate 3D MRI-to-CT synthesis.

In this work, we introduce Parallel Swin Transformer-Enhanced Med2Transformer, a novel 3D network that enhances the encoder and bottleneck with parallel Swin Transformer blocks combined with convolutional layers of different dilation rates. This design jointly models local and global features through multi-scale shifted-window attention and hierarchical feature fusion, improving geometric and dosimetric consistency. The proposed method is evaluated on public and clinical datasets and benchmarked against the top SynthRAD2023 challenge solutions (transformer-based MTT-Net [5] and nnU-Net based [6]) and our prior Med-DDPM [7] diffusion model, demonstrating superior image fidelity and clinically acceptable dose accuracy.

## 2. METHODS

### 2.1. Model Overview

The proposed Parallel Swin Transformer-Enhanced Med2Transformer is a 3D encoder-decoder network for MRI-to-CT synthesis. Built upon the MTT-Net [5] backbone, the model incorporates parallel Swin Transformer blocks combined with convolutions of different dilation rates in the encoder and bottleneck. This enhancement enables simultaneous modeling of fine local structures and global anatomical context, improving image similarity and geometric consistency in sCT generation. An overview of the architecture is presented in Fig. 1.

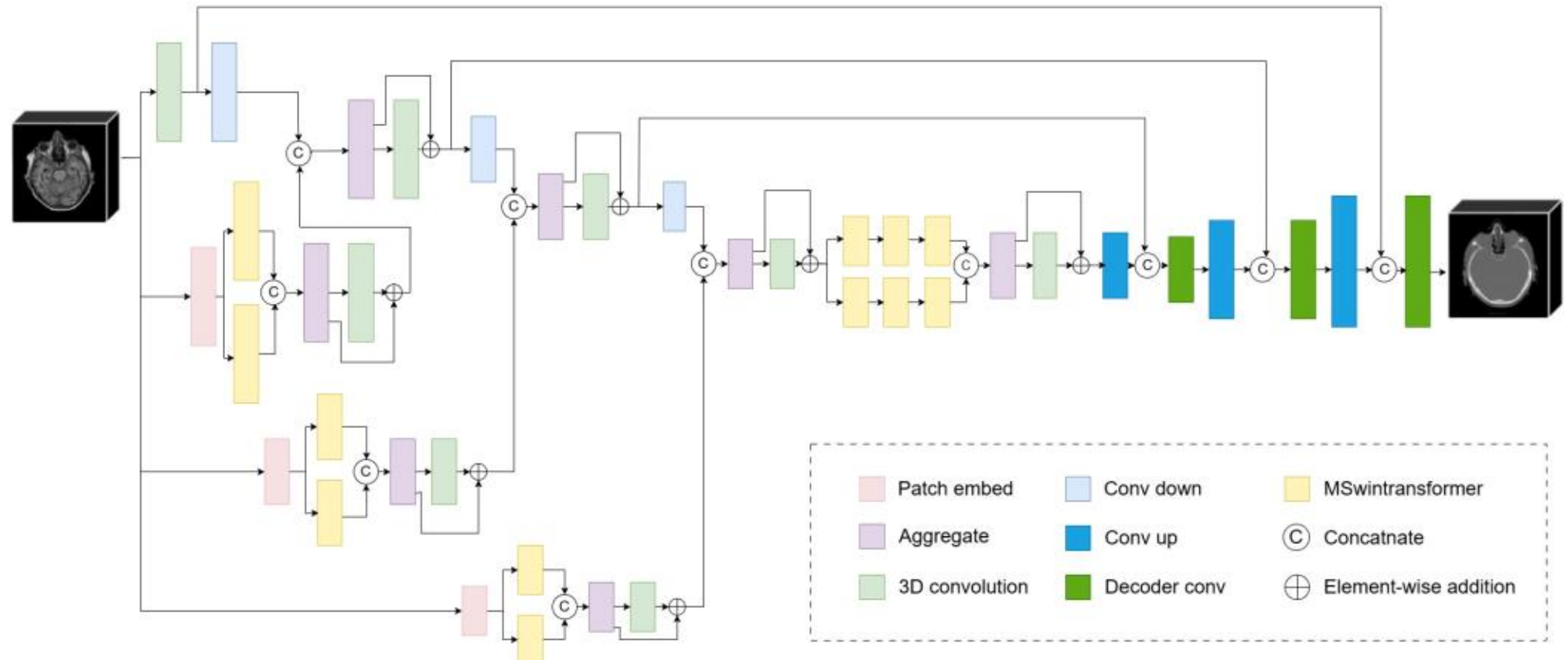

**Fig. 1.** Overview of the proposed Parallel Swin Transformer-Enhanced Med2Transformer architecture. The encoder and bottleneck are enhanced with parallel Swin Transformer blocks integrated with convolutional layers of different dilation rates, enabling joint modeling of local details and global context. Multi-scale feature fusion ensures anatomically consistent MRI-to-CT synthesis.

### 2.2. Parallel Swin Transformer Enhancement

Each encoder stage consists of a convolutional stream and a parallel Swin Transformer branch. The convolutional stream extracts local features using $3 \times 3 \times 3$ kernels with *Instance Normalization* and *ReLU* activation. The transformer branch models long-range dependencies via multi-scale shifted-window self-attention. To diversify the receptive field, convolutional layers with different dilation rates ($d \in \{1,2\}$) are applied before each transformer block, enhancing contextual awareness without increasing parameter count. At layer $\ell$, features from both branches are fused through channel-wise concatenation followed by residual projection:

$$f^{(\ell)} = \mathrm{Conv}_{1\times1\times1}([f_c^{(\ell)}, f_t^{(\ell)}]) + f_c^{(\ell)},$$

where $f_c^{(\ell)}$ and $f_t^{(\ell)}$ denote convolutional and transformer features, respectively. This parallel design strengthens anatomical representation and stabilizes training across heterogeneous MRI domains.

### 2.3. Bottleneck and Decoder

The bottleneck extends the parallel configuration with two Swin Transformer blocks using different dilation settings to capture deep semantic context. The decoder progressively upsamples features via trilinear interpolation followed by $3 \times 3 \times 3$ convolutions. Skip connections from the encoder preserve high-resolution spatial information and improve geometric alignment. The final layer applies a $1 \times 1 \times 1$ convolution with *tanh* activation to generate an HU-normalized sCT volume.

### 2.4. Adversarial Framework

The generator is trained adversarially using a Wave3D Discriminator, which integrates 3D convolutional filtering with wavelet-based frequency decomposition to emphasize fine structural details. The discriminator enforces CT realism through adversarial supervision, while spatial and perceptual losses preserve voxel-level and anatomical accuracy. The overall training objective is a composite loss function defined in Eq. (1).

### 2.5. Loss Functions

The generator is optimized using a weighted combination of adversarial, reconstruction, and perceptual losses:

$$\mathcal{L}_{\text{total}} = \lambda_{\text{GAN}}\mathcal{L}_{\text{GAN}} + \lambda_{L1}\mathcal{L}_{L1} + \lambda_{\text{perc}}\mathcal{L}_{\text{perc}} \ , \qquad (1)$$

with $\lambda_{\text{GAN}} = 1$, $\lambda_{L1} = 20$, and $\lambda_{\text{perc}} = 1$. The adversarial term $\mathcal{L}_{\text{GAN}}$ adopts a BCE formulation to promote realistic CT texture against the Wave3D Discriminator. The reconstruction loss $\mathcal{L}_{L1}$ minimizes voxel-wise intensity differences between synthesized and reference CT volumes. The perceptual loss $\mathcal{L}_{\text{perc}}$ computes the $L_1$ distance between *VGG-19* feature maps of generated and ground-truth slices, encouraging structural and textural fidelity. This composite objective balances voxel accuracy, structural consistency, and perceptual realism during training.

## 3. EXPERIMENTS AND RESULTS

Med2Transformer was trained using the Adam optimizer ($\beta_1 = 0.5$, $\beta_2 = 0.999$) with a maximum learning rate of $2 \times 10^{-4}$, decayed over 100 epochs. Training used 3D MRI patches of size $128 \times 128 \times 48$ with a batch size of 12. The adversarial learning framework optimized the generator and discriminator using a weighted L1 loss ($\lambda = 20$), a vanilla GAN loss, and a VGG perceptual loss. Experiments on the public dataset were conducted on an NVIDIA Cloud GDX node with eight H100 GPUs, while the private clinical dataset was trained on a single A100 GPU under deterministic PyTorch settings. Full volume sCT inference employed a sliding window strategy with overlap aware averaging.

**Datasets:** Experiments were performed on two datasets.
(1) The SynthRAD 2023 public dataset, containing paired MRI-CT scans of the brain and pelvis. For each anatomical region, the dataset provides 139 training, 5 validation, and 36 testing pairs (360 cases in total).
(2) An IRB-approved NTUH clinical head-and-neck dataset with 1,439 paired MRI-CT cases, of which 1,362 were allocated to training/validation and 77 to testing.
All images were rigidly aligned, resampled to 1 $\text{mm}^3$, and intensity normalized to $[-1, 1]$.

**Baselines:** We compared our method with three representative baselines: MTT-Net [5], a transformer based model and 1st place winner of the SynthRAD 2023 Challenge; nnU-Net [6], the 2nd place winner; and Med-DDPM [7], our prior diffusion based model. All methods were trained using identical data preprocessing, patch extraction, and evaluation protocols. The original nnU-Net and Med-DDPM outputs showed degraded quantitative performance due to pixel intensity inconsistencies. Therefore, additional post-processing was applied to align intensity ranges, improving quantitative performance.

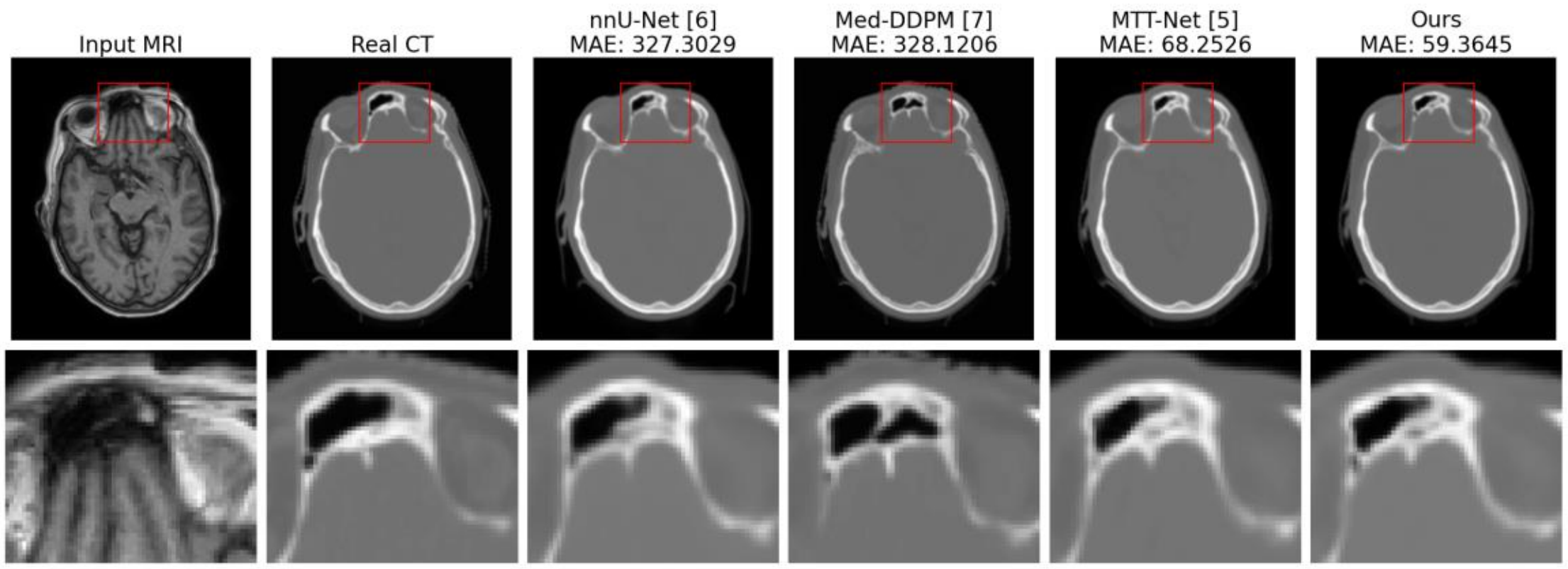

**Fig. 2.** Axial center-cut slices compare the input brain MRI, reference CT, and synthetic CTs generated by nnU-Net, Med-DDPM, MTT-Net, and the proposed Med2Transformer, with corresponding MAE values. Zoomed-in views of the frontal sinus region highlight differences in cortical and trabecular bone depiction. Med2Transformer yields sharper osseous boundaries and more anatomically consistent morphology, producing the closest correspondence to the reference CT.

**Table 1.** Quantitative image-similarity evaluation of brain and pelvis sCT

| Model | Brain | | | Pelvis | | |
|---|---|---|---|---|---|---|
| | MAE (HU)↓ | SSIM↑ | PSNR (dB)↑ | MAE (HU)↓ | SSIM↑ | PSNR (dB)↑ |
| nnU-Net [6] | 91.4654 | 0.8425 | 27.9934 | 103.0749 | **0.8455** | 27.1852 |
| Med-DDPM [7] | 91.3063 | **0.9332** | **30.4090** | 98.1615 | 0.7969 | 27.0672 |
| MTT-Net [5] | 80.6294 | 0.8536 | 27.7128 | 78.4184 | 0.8255 | 27.3311 |
| Ours | **71.9605** | 0.8701 | 28.0453 | **72.0720** | 0.8401 | **27.7139** |

**Table 2.** Dice scores for geometric consistency of brain and pelvis sCT by selected anatomical structures

| Model | Brain (Mean Dice↑) | | | | | Pelvis (Mean Dice↑) | | | | |
|---|---|---|---|---|---|---|---|---|---|---|
| | GM | CSF | Bone | ST | Mean | Hip-L | Hip-R | Sacrum | Bladder | Mean |
| nnU-Net [6] | 0.90 | 0.83 | 0.88 | 0.87 | 0.87 | 0.85 | 0.86 | 0.83 | 0.63 | 0.79 |
| Med-DDPM [7] | 0.91 | **0.88** | 0.90 | 0.89 | 0.90 | 0.87 | 0.89 | 0.84 | 0.63 | 0.81 |
| MTT-Net [5] | 0.91 | 0.85 | 0.89 | 0.90 | 0.89 | 0.85 | 0.87 | 0.79 | 0.64 | 0.79 |
| Ours | **0.92** | 0.86 | **0.90** | **0.90** | **0.90** | **0.88** | **0.89** | **0.85** | **0.67** | **0.82** |

**Evaluation Metrics:** Quantitative image quality was assessed using MAE, SSIM, and PSNR within body masks for brain and pelvis CT synthesis. Geometric consistency was evaluated using Dice scores by comparing segmentation masks from synthetic CTs with those from real CTs. CTseg [8] was used to segment head structures (gray matter, cerebrospinal fluid, bone, soft tissue), while TotalSegmentator [9] was used for pelvis structures (hips, sacrum, bladder). Preliminary dose evaluation was additionally performed by NTUH expert using the CyberKnife system.

**Quantitative Results:** Table 1 summarizes the quantitative evaluation on brain and pelvis sCT synthesis. Our method achieves the lowest MAE for both anatomies (72.0 HU for brain and 72.1 HU for pelvis), indicating improved voxel-wise accuracy compared to all baselines. It also demonstrates strong structural similarity, with SSIM values of 0.870 for brain and 0.840 for pelvis, and achieves the highest PSNR on the pelvis dataset (27.7 dB). While Med-DDPM reports higher SSIM (0.933) and PSNR (30.4 dB) on the brain dataset, our method demonstrates more consistent performance across anatomies, combining reduced intensity error with competitive structural fidelity.

**Geometric Consistency:** Table 2 reports segmentation accuracy for selected anatomical structures in the brain and pelvis. Our method achieved the highest mean Dice scores, 0.90 for brain and 0.82 for pelvis. Compared with MTT-Net, our model showed improvements on key structures, including GM (0.92 vs. 0.91), Bone (0.90 vs. 0.89), and Hip-L/R (0.88/0.89 vs. 0.85/0.87). These results indicate better geometric alignment and boundary consistency.

**Qualitative Evaluation:** Across brain and pelvis datasets, the proposed method generates sCT with enhanced cortical bone delineation and more consistent soft-tissue contrast compared with baseline models. It also preserves anatomical boundaries more accurately in high-gradient regions. Fig. 2 illustrates an example in which Med2Transformer provides clearer osseous structures and closer correspondence to the reference CT than nnU-Net, Med-DDPM, and MTT-Net.

**Dosimetric Evaluation:** A CyberKnife-based dose analysis was conducted on a representative head-and-neck case selected from the NTUH cohort and independently reviewed by an NTUH radiation oncology expert. The synthetic CT (sCT)-based plan showed close agreement with the reference CT in target coverage and organ-at-risk (OAR) doses. For the primary target volume ($TV_1$), the sCT exhibited a 1.69% mean-dose deviation and a 2.23% reduction in coverage relative to the planning CT. Across 20 anatomical structures, the average mean-dose difference was 3.8%, indicating consistent dose behavior. These results demonstrate the feasibility of sCT-based dose computation and motivate further large-scale clinical validation.

## 5. CONCLUSION

We presented the Parallel Swin Transformer-Enhanced Med2Transformer, a 3D framework for MRI-to-CT synthesis aimed at enabling MRI-only radiotherapy planning. The model enhances the encoder and bottleneck with parallel Swin Transformer blocks and convolutions of varying dilation rates to jointly capture local details and global anatomical context.

Experiments on public (SynthRAD2023) and clinical (NTUH) datasets demonstrated improved image similarity and geometric consistency compared with baseline methods. Quantitative dose evaluation showed a 1.69% target-dose deviation, supporting the feasibility of sCT-based planning. Future work will benchmark against leading methods from the recent SynthRAD2025 challenge and pursue large-scale clinical validation to further support the reliability of MRI-only workflows.

## 6. COMPLIANCE WITH ETHICAL STANDARDS

The SynthRAD2023 dataset was obtained under the CC BY-NC 4.0 license and required no additional ethical approval. The head and neck MRI–CT dataset from National Taiwan University Hospital was used with Institutional Review Board approval (IRB No. 202403019RINA).

## 7. ACKNOWLEDGMENTS

This work was supported in part by the National Science and Technology Council of Taiwan under Grants 114-2221-E-011-058-MY3 and 114-2634-F-011-002-MBK. The authors acknowledge NVIDIA Corporation for providing computational resources through the National Taiwan University Artificial Intelligence and High Performance Computing Application Research Project (NVIDIA Taipei-1)

## 8. REFERENCES

[1] M. A. Bahloul *et al*., "Advancements in synthetic CT generation from MRI: A review of techniques and trends in radiation therapy planning," *J. Appl. Clin. Med. Phys.*, vol. 25, no. 11, Art. e14499, 2024.

[2] R. J. Goodburn *et al*., "The future of MRI in radiation therapy: Challenges and opportunities for the MR community," *Magn. Reson. Med.*, vol. 88, no. 6, pp. 2592-2608, 2022.

[3] F. Villegas *et al*., "Challenges and opportunities in the development and clinical implementation of artificial-intelligence-based synthetic computed tomography for magnetic-resonance-only radiotherapy," *Radiother. Oncol.*, Art. 110387, 2024.

[4] M. K. Sherwani and S. Gopalakrishnan, "A systematic literature review: Deep learning techniques for synthetic medical image generation and their applications in radiotherapy," *Front. Radiol.*, vol. 4, Art. 1385742, 2024.

[5] L. Zhong *et al*., "Multi-scale tokens-aware transformer network for multi-region and multi-sequence MR-to-CT synthesis in a single model," *IEEE Trans. Med. Imaging*, vol. 43, no. 2, pp. 794-806, Feb. 2024, doi: 10.1109/TMI.2023.3321064.

[6] A. Thummerer *et al*., "SynthRAD2023 Grand Challenge dataset: Generating synthetic CT for radiotherapy," *Med. Phys.*, vol. 50, no. 7, pp. 4664-4674, 2023.

[7] Z. Dorjsembe, H.-K. Pao, S. Odonchimed, and F. Xiao, "Conditional diffusion models for semantic 3D brain MRI synthesis," *IEEE J. Biomed. Health Inform.*, 2024.

[8] WCHN, "CTseg: Automated CT segmentation toolkit." GitHub repository. Available: https://github.com/WCHN/CTseg

[9] J. Wasserthal *et al*., "TotalSegmentator: Robust segmentation of 104 anatomical structures in CT." GitHub repository. Available: https://github.com/wasserth/TotalSegmentator